\documentclass[conference, letterpaper, 10 pt]{IEEEtran}

\IEEEoverridecommandlockouts                              

\usepackage{cite}
\usepackage{amsmath,amssymb,amsfonts}
\usepackage{graphicx}
\usepackage{textcomp}

\usepackage{algorithm}
\usepackage{algpseudocode} 
\usepackage{xcolor}
\usepackage{multicol}
\usepackage{multirow}
\usepackage{subcaption}
\usepackage[export]{adjustbox} 
\providecommand{\keywords}[1]{\textbf{\textit{Index terms---}} #1}
\usepackage{booktabs}
\newcommand{\ra}[1]{\renewcommand{\arraystretch}{#1}}

\usepackage[colorlinks=true,allcolors=red]{hyperref}

\title{\LARGE\bf
Early Stopping Criteria for Training Generative Adversarial Networks in Biomedical Imaging}

\author{Muhammad Muneeb Saad$^{1}$, Mubashir Husain Rehmani$^{2}$ and Ruairi O'Reilly$^{3}$

\thanks{$^{1,2,3}$MM. Saad, MH. Rehmani and R. O'Reilly are with Munster Technological University Cork, Ireland. *Corresponding author to email {\tt\small muhammad.saad@mycit.ie.}
} \newline
\thanks{979-8-3503-5298-6/24/\$31.00 ©2024 IEEE}}

\begin{document}

\maketitle
\thispagestyle{empty}
\pagestyle{plain}

\begin{abstract}
Generative Adversarial Networks (GANs) have high computational costs to train their complex architectures. Throughout the training process, GANs' output is analyzed qualitatively based on the loss and synthetic images' diversity and quality. Based on this qualitative analysis, training is manually halted once the desired synthetic images are generated. By utilizing an early stopping criterion, the computational cost and dependence on manual oversight can be reduced yet impacted by training problems such as mode collapse, non-convergence, and instability. This is particularly prevalent in biomedical imagery, where training problems degrade the diversity and quality of synthetic images, and the high computational cost associated with training makes complex architectures increasingly inaccessible. This work proposes a novel early stopping criteria to quantitatively detect training problems, halt training, and reduce the computational costs associated with synthesizing biomedical images. Firstly, the range of generator and discriminator loss values is investigated to assess whether mode collapse, non-convergence, and instability occur sequentially, concurrently, or interchangeably throughout the training of GANs. Secondly, utilizing these occurrences in conjunction with the Mean Structural Similarity Index (MS-SSIM) and Fréchet Inception Distance (FID) scores of synthetic images forms the basis of the proposed early stopping criteria. This work helps identify the occurrence of training problems in GANs using low-resource computational cost and reduces training time to generate diversified and high-quality synthetic images.
\end{abstract}


\section{INTRODUCTION}
Increasingly complex GAN architectures introduce significant computational challenges to train and generate synthetic images~\cite{ji2023alpha}. These challenges arise from the GAN architecture and size of datasets. One of the primary limitations is the computational cost associated with processing large volumes of high-resolution images through multiple training iterations \cite{dumont2021overcoming}.

GANs are developed with the generator model for generating synthetic images and the discriminator model for distinguishing synthetic images from real images. These two models are trained using an adversarial approach where the generator aims to create realistic images to mislead the discriminator and the discriminator aims to classify real and synthetic images accurately \cite{karnewar2020msg}.

Popular GANs are designed with complex architectures as indicated in Table \ref{tab:complexityofgans}. Training these GANs necessitates optimizing complex objective functions requiring substantial computing and memory resources. Generating desirable synthetic images often requires running multiple iterations of training epochs, further enhancing the computational cost~\cite{lin2021anycost}.

\begin{table*}[htp!]
\centering
\ra{1.1}
\caption{\textbf{An overview of GAN architectures detailing factors contributing to the complexity and computational cost during a single training cycle of GANs. The memory requirements for all GANs refer to the training from scratch.}}
\begin{tabular}{p{1.5cm}p{1.5cm}p{1.7cm}p{1.7cm}p{2.2cm}p{0.8cm}p{3cm}p{2cm}}
\toprule
\textbf{GANs} & \textbf{Generator} & \textbf{Discriminator} & \textbf{Loss} & \textbf{TS} & \textbf{TI} & \textbf{AC} & \textbf{CPU/GPU} \\
\midrule

MSGGAN~\cite{karnewar2020msg} & Multi-scale gradient layers & Multi-scale gradient layers & WGAN-GP & Multi-scale gradient using PG & 100K & Hypersphere normalization & Cluster (8-GPUs) \\
DCGAN~\cite{radford2015funsupervised} & Deconvolutional layers & Convolutional layers & BCE & MinMax & 1-5K & N/A & CPU/GPU \\
StyleGAN~\cite{karras2019style} & Neural style transfer & PG Layers & WGAN-GP & PG and Mixing R1 & 100K & Mixing R1, BU, 8-layers MLP MN and IN, Noise & Cluster (8-GPUs) \\
StyleGAN2~\cite{karras2020training} & Similar to Style-GAN & Augmented Layers & Non-saturating & Adaptive discrminator augmentation & 500K & Balanced Consistency R1 and Mixed Precision & Cluster (8-GPUs) \\
StyleGAN3~\cite{karras2021alias} & StyleGAN2 with Fourier transform layers & StyleGAN2 & Similar to StyleGAN2 & Similar to StyleGAN2 & 500K & Flexible layers specifications using Fourier features and rotation equivariance in training & Cluster (8-GPUs) \\
TransGAN~\cite{jiang2021transgan} & Transformer with PG layers & Multi-scale layers with grid self-attention & WGAN-GP & Adopting relative position encoding & 400K & Data augmentation and modifying layer normalization & Cluster (16-GPUs) \\
StyleSwin~\cite{zhang2022styleswin} & Swin transformer with neural style & Wavelet layers & Total variation annealing & Two-timescales training updates & 25.6M & Double attention & Cluster (8-GPUs) \\

\bottomrule
\multicolumn{8}{l}{AC: Additional Components; BU: Bilinear Upsampling; Dis:Discriminator; Gen:Generator; IN: Instance Normalization; MN: Mapping-} \\
\multicolumn{8}{l}{Network; PG: Progressive Growing; R1: Regularization; TS:Training Scheme; TI:Training Iterations} \\
\end{tabular}
\label{tab:complexityofgans}
\end{table*}
GANs are widely utilized for synthesizing images in the application domains under consideration. When a GAN fails to produce such images, it requires additional training cycles and modifications to the architecture. Three common problems contributing to a GAN's failure to create synthetic images are mode collapse, non-convergence, and instability. Mode collapse occurs when a GAN generates synthetic images with similar feature distributions from input images having distinct feature distributions. Non-convergence relates to a GAN in an imbalanced training state as indicated by a deviation from the Nash equilibrium. Instability in GAN is indicative of a vanishing gradient problem. The generator and discriminator losses throughout GANs training cycles are evaluated to identify these problems \cite{abdusalomov2023evaluating}. However, identifying each problem individually during training is challenging.

The Multi-scale Structural Similarity Index Measure (MS-SSIM) and the Fréchet Inception Distance (FID) are considered unified metrics to quantify the diversity and quality of synthetic images. These metrics are widely used as MS-SSIM quantifies the diversity between images using perceptual similarity of features, and FID quantifies the quality of synthetic images by evaluating the distance between real and synthetic images \cite{odena2017conditional}~\cite{segal2021evaluating}.

Generating synthetic images requires criteria for acceptable ranges of diversity and quality compared to real images. However, GANs lack focus on these acceptable ranges for diversity and quality of images~\cite{saad2023self}~\cite{zhao2021improved}. The loss values of the generator and discriminator also have no consensus to identify acceptable ranges of diversity and quality of images, initiating an additional problem for training GANs with unknown training cycles that raise computational costs.

In the biomedical imaging domain, the qualitative assessment of synthetic images often requires the expertise of medical professionals, leading to additional costs \cite{dash2023review}. While MS-SSIM and FID metrics are commonly used for quantitative assessment, there is a pressing need for a consensus on acceptable ranges of diversity and quality for synthetic images. This lack of agreement hampers progress in the field, underscoring the need for the proposed work. As a result, the training of GANs can be seen as a bottleneck in the field of biomedical imaging, where computational resources are often either underutilized or overutilized. This inefficiency stems from a lack of informed acceptable ranges of diversity and quality for synthetic images. By exploring early stopping criteria that combine loss values with these acceptable ranges, we can establish a more efficient training process, thereby reducing the utilization of computational resources and increasing the accessibility of GANs' within the field of biomedical imaging.

The need to quantify the range of loss values to identify mode collapse, non-convergence, and instability problems in GANs is pressing. The dynamic changes in loss behavior during training, based on GAN characteristics and image type, pose significant technical challenges. Therefore, investigating the range of generator and discriminator loss values that can effectively identify GAN training problems is warranted.

Early stopping criteria are commonly adopted using training and validation losses in image classification tasks \cite{ji2023alpha}. However, a comparison of Table \ref{evaluatingtrainingproblems} and Table \ref{earlystoppliterature} highlights that GAN-based studies do not routinely utilize early stopping criteria. While those that do, Table \ref{earlystoppliterature}, rely on loss and qualitative analysis for early stopping during training. Again, there is no consensus on adopting early stopping to achieve high-quality synthetic images with efficient GAN training.

\subsection{Research Contributions}

\begin{itemize}
    \item This work explores the range of generator and discriminator loss values to identify mode collapse, non-convergence, and instability problems. Furthermore, it seeks to examine the relationship among these issues to determine if they occur sequentially, concurrently, or interchangeably during the training of GANs.
    \item Investigate early stopping criteria based on loss values combined with the MS-SSIM and FID scores to enable better identification of the training problems via diversity and quality of synthetic images while reducing training time and usage of computational resources during the training cycles of GANs.
\end{itemize}

\section{RELATED WORK}
In related work, it's noted that while generator and discriminator losses are commonly used to identify training problems in GANs, relying solely on loss values may not accurately pinpoint these issues. Additionally, existing studies (See Table \ref{evaluatingtrainingproblems}) lack specificity regarding the range of these loss values and fail to report the variance in generator and discriminator losses across each problem, which presents a gap in the literature.

Consequently, many studies employ a qualitative assessment of loss values and synthetic images evaluating the diversity and quality, enabling a challenge for non-experts. No correlation is being discussed between loss values, diversity, and the quality of synthetic images. 
\begin{table}[htp!]
\centering
\caption{\textbf{Overview of Studies Using Loss and Qualitative Assessment of Synthetic Images to Identify GAN Training Problems in Biomedical and Non-Biomedical Images.}}
\begin{tabular}{p{0.8cm}p{1.5cm}p{0.8cm}p{0.8cm}p{0.2cm}p{0.2cm}p{0.2cm}p{0.2cm}}
\toprule
\textbf{Domain} & \textbf{GANs} & \textbf{Imagery} & \textbf{Loss} & \textbf{Q.A} & \textbf{M.C} & \textbf{N.C} & \textbf{Inst.} \\
\midrule

\multirow{4}{*}{General} & $\alpha$EGAN~\cite{ji2023alpha} & CelebA & Energy Distance & \checkmark & \checkmark & n/a & n/a \\
& DCGAN~\cite{heusel2017gans} & CelebA & BCE & n/a & \checkmark & n/a & n/a \\
& DCGAN~\cite{zhao2021improved} & CelebA & Non-Saturating & \checkmark & \checkmark & n/a & n/a \\
& StyleGAN~\cite{beckham2022overcoming} & EMNST & Info-Max & \checkmark & \checkmark & n/a & n/a \\
&&&&&& \\
\multirow{8}{*}{Biomed} & progGAN~\cite{dumont2021overcoming} & X-ray & Hinge-WGAN & \checkmark & n/a & n/a & \checkmark \\
& PGGAN~\cite{segal2021evaluating} & X-ray & WGAN-GP & \checkmark & n/a & n/a & n/a \\
& DCGAN~\cite{belmekki2022empirical} & X-ray & BCE & \checkmark & \checkmark & n/a & \checkmark \\
& CGAN~\cite{al2023usability} & MRI & Focal Tversky & \checkmark & \checkmark & n/a & n/a \\
& GAN~\cite{valvano2021stop} & MRI & CE & n/a & n/a & \checkmark & n/a \\
& DCGAN~\cite{abdusalomov2023evaluating} & Ultra. & BCE & \checkmark & \checkmark & n/a & n/a \\
& DCGAN~\cite{saad2023assessing} & X-ray & BCE & \checkmark & \checkmark & \checkmark & n/a \\
& DCGAN~\cite{saad2022addressing} & X-ray & BCE & \checkmark & \checkmark & n/a & n/a \\
& MSGGAN~\cite{saad2023self} & X-ray & RHinge & \checkmark & n/a & n/a & \checkmark \\

\bottomrule
\multicolumn{8}{l}{BCE: Binary Cross-entropy; Biomed: Biomedical; Inst: Instability} \\
\multicolumn{8}{l}{M.C: Mode Collapse; N.C: Non-convergence; Q.A: Qualitative} \\
\multicolumn{8}{l}{Assessment of Synthetic images; Ref: Reference; Ultra: Ultrasound} \\

\end{tabular}
\label{evaluatingtrainingproblems}
\end{table}

A limited number of studies \cite{ji2023alpha} \cite{yang2022generalization} propose early stopping criteria as a separate function for training GANs. Authors in \cite{ji2023alpha} emphasize only GAN loss without considering quantitative measures of diversity and quality of images. Similarly, theoretical analysis is proposed for early stopping in GANs without providing any quantitative measures for loss, diversity, or quality of synthetic images \cite{yang2022generalization}. Table \ref{earlystoppliterature} indicates the existing studies that use early stopping points in training GANs either qualitatively based on loss and synthetic images or quantitatively with FID scores. Qualitative and quantitative methods reported in Table \ref{earlystoppliterature} indicate a need for more information for early stopping heuristics in GANs. 

Consequently, assessing synthetic image diversity and quality through loss values and qualitative assessment for early stopping criteria poses challenges for non-experts. Hence, early stopping criteria combining loss values, MS-SSIM, and FID scores to quantitatively evaluate synthetic image diversity and quality while reducing the training time and computational costs need to be explored.
\begin{table}[htp!]
\centering
\caption{\textbf{Existing studies utilizing early stopping criteria in training GANs for generating synthetic images.}}

\begin{tabular}{p{1.6cm}p{0.3cm}p{2.6cm}p{2.5cm}}
\toprule
\textbf{GANs (Im)} & \textbf{Loss} & \textbf{FID} & \textbf{Early Stopping} \\
\midrule

$\alpha$ (FI)~\cite{ji2023alpha} & ED & 114.6 (200k iterations) & Loss and synthetic images visualized qualitatively \\
prog (X)~\cite{dumont2021overcoming} & HW & 3.0 (missing) & FID stops improving \\
DC (FI)~\cite{heusel2017gans} & BCE & 12.5 (225k iterations) & FID stops improving \\
DC (FI)~\cite{zhao2021improved} & NS & 15.43 (100k iterations) & No heuristics \\
St (HD)~\cite{beckham2022overcoming} & IM & 7.5 (1k epochs) & FID stops improving \\
DC (X)~\cite{belmekki2022empirical} & BCE & 100.58  (400 epochs) & Synthetic images visualized qualitatively \\
DC (US)~\cite{abdusalomov2023evaluating} & BCE & 34.41 (1k epochs) & Synthetic images visualized qualitatively \\

\bottomrule
\multicolumn{4}{l}{$\alpha$:$\alpha$EGAN; BCE:Binary Cross-entropy; DC:DCGAN; ED:Energy} \\
\multicolumn{4}{l}{Distance; FI:Face Images; HD:Handwritten Digits; HW:Hinge-WGAN} \\
\multicolumn{4}{l}{Im:Images; IM:Info-Max; K:1000; NS:Non-Saturating; prog:prog-GAN} \\
\multicolumn{4}{l}{St:StyleGAN; US:Ultrasound; X:X-ray} \\

\end{tabular}
\label{earlystoppliterature}
\end{table}

\section{METHODOLOGY}

\subsection{Dataset}
In this work, a publicly available dataset comprising chest X-ray images of 10,192 healthy cases as the majority class and 3616 coronavirus (COVID-19) as the minority class \cite{rahman2021exploring} was accessed. The dataset was divided into 80\% training and 20\% testing sets. Subsequently, the training set contains 8153 healthy and 2893 COVID-19 images while the test set contains 2039 healthy and 723 COVID-19 images. The COVID-19 images from the training set were selected for generating synthetic images using GANs. These images were resized to 128x128 resolution for training GANs.

\subsection{GAN Architectures}
In this work, a DCGAN was implemented due to requiring low-resource computes \cite{dash2023review} following the architecture detailed in \cite{saad2022addressing}. The DCGAN was trained on 128x128 resolution images. Due to computational constraints, hyperparameters including a batch size of 4, and 1000 training epochs were selected. The constant learning rates (LR) of 0.001, 0.002, and 0.003 were utilized in different training cycles to identify the training problems in DCGAN. To optimize the training process, scheduled LR for the generator and discriminator models were employed. These LR were initialized at 0.0001, allowing for gradual adjustments to enhance convergence and stability throughout the training of GANs. The DCGAN was trained using binary cross-entropy (BCE) loss. The generator and discriminator losses of DCGAN \cite{belmekki2022empirical} are defined in Eqs. \ref{loss_generator} and \ref{loss_discrminator}. 
\begin{equation}
\vspace{-1cm}
\mathcal{L}_{\text{G}} = -\mathbb{E}_{z\sim p_{\text{z}}(z)}[\log D(G(z))]\label{loss_generator}
\vspace{0.3cm}
\end{equation}

\begin{equation}
\vspace{0.1cm}
\mathcal{L}_{\text{D}} = -\mathbb{E}_{x\sim p_{\text{data}}(x)}[\log D(x)] - \mathbb{E}_{z\sim p_{\text{z}}(z)}[\log (1 - D(G(z)))]\label{loss_discrminator}
\end{equation}
In Eqs. \ref{loss_generator} and \ref{loss_discrminator}, ${\text{G}}(z)$ represents the generator's output with input ${\text{z}}$, ${\text{D}}(x)$ represents the discriminator's output with real input images ${\text{x}}$, $p_{\text{data}}(x)$ represents the distribution of real data, $p_{\text{z}}(z)$ represents the distribution of random input ${\text{z}}$, and $\mathbb{E}$ denotes the expected value over the respective distributions.

The Multi-Scale Gradient GAN (MSG-GAN) was also implemented utilizing relativistic hinge loss as detailed in \cite{saad2023self}, as a comparator. The MSG-GAN was trained with the same batch size, 128x128 resolution, and number of training images as the DCGAN. The LR of 0.003 was used for both the generator and discriminator models, as suggested in \cite{saad2023self}.

\subsection{Identifying Training Problems via Loss Values}
The correlation between generator and discriminator loss values is assessed according to the quality of synthetic images. The binary cross-entropy (BCE) loss in DCGAN is used to quantify the range of loss values of the generator and discriminator. These ranges are used to identify the training problems in DCGAN. Similarly, the relativistic hinge loss in MSG-GAN is used to identify the training problems. 

\subsection{Identifying Training Problems via Diversity and Quality}
MS-SSIM scores are used to assess the occurrence of the mode collapse problem. These scores quantify the diversity of the synthetic images generated. In this work, MS-SSIM is computed dynamically every 50 epochs by selecting 50 image pairs (100 image samples) randomly from the datasets (training, test, and synthetic), due to the availability of low computational resources. The mean MS-SSIM score for each dataset provides a comparative analysis of the diversity of synthetic images with training and test datasets. A lower MS-SSIM score indicates a higher diversity of synthetic images.

FID scores are used to assess the occurrence of non-convergence and instability problems. It measures the quality of synthetic images. FID is also calculated dynamically every 50 epochs by selecting 100 image samples randomly for train-to-train, train-to-test, and train-to-synthetic datasets. The mean FID scores are subsequently reported. A lower FID score indicates a higher quality of synthetic images.

\subsection{Early Stopping Criteria}
The proposed early-stopping criterion combines the loss values of the generator and discriminator as well as the MS-SSIM and FID scores of synthetic images. A GAN's training is terminated under the following conditions:
\begin{enumerate}
    \item Loss of the generator and discriminator depicts training problems (sharp increase/decrease, oscillating values, and consistent values).
    \item MS-SSIM score of synthetic images becomes less than or equal to the MS-SSIM thresholds (MS-SSIM (train images) and MS-SSIM (test images)) scores.
    \item FID score of train-synthetic images becomes less than or equal to the FID thresholds (FID (train-train images) and FID (train-test images)) scores.
\end{enumerate}
The early stopping criterion is implemented in two steps. In the first step, the loss values were used as the early stopping criterion for training DCGAN. In the second step, a combined metric for loss, MS-SSIM, and FID scores were used as the early stopping criterion for GAN training. The patience epochs 50, 100, and 200 were used in the early stopping criterion to allow GAN training for improvement. A comparative analysis of early stopping criteria is reported in Section \ref{results_discussion}. The proposed early stopping criterion in a GAN training is reported in Algorithm \ref{alg:training}.

\begin{algorithm}[htp!]
\caption{Early stopping based on loss, MS-SSIM, and FID in training GANs.}
\label{alg:training}
\begin{algorithmic}[1]
    \Require Train and Test image datasets, Generator (G), Discriminator (D), Max epochs (max\_epochs), MS-SSIM Train threshold (MS-SSIM\_Th1), MS-SSIM Test threshold (MS-SSIM\_Th2), MS-SSIM Synthetic (MS-SSIM), FID Train-Train threshold (FID\_Th1), FID Train-Test threshold (FID\_Th2), FID Train-Synthetic (FID), Patience Epochs (patience).
    \Ensure Trained G and D models:
    \State Set best\_MS-SSIM\_score to MS-SSIM\_Th1 and MS-SSIM\_Th2, and best\_FID\_score to FID\_Th1 and FID\_Th2.
    \State Set patience epochs, no\_improvement\_count, and loss\_problem\_count.
    \State Initialize list to store recent loss values: recent\_loss\_G and recent\_loss\_D.
    \ForAll{$\text{epoch} = 1$ \textbf{to} max\_epochs}
        \State Train D and G networks for one epoch.
        \State Record the loss values of G and D for this epoch.
        \State Append current loss values to recent\_loss\_G and recent\_loss\_D.
        \If{analyze\_loss\_patterns(recent\_loss\_G, recent\_loss\_D)}
            \State Increment loss\_problem\_count by 1.
        \Else
            \State Reset loss\_problem\_count to 0.
        \EndIf
        \If{loss\_problem\_count $\geq$ patience}
            \State Terminate Training: Early stopping due to training problems detected in loss values for patience epochs.
            \State \textbf{break}
        \EndIf
        \If{$\text{epoch mod 50} == 0$}
            \If{constant loss values indicating no improvement are detected}
                \State Calculate MS-SSIM\_Th1, MS-SSIM\_Th2, MS-SSIM,
                FID\_Th1, FID\_Th2, and FID scores.
                \If{MS-SSIM score $\leq$ best\_MS-SSIM\_score and FID score $\leq$ best\_FID\_score}
                    \State Update best\_MS-SSIM\_score to MS-SSIM score and best\_FID\_score to FID score.
                    \State Reset no\_improvement\_count to 0.
                \Else
                    \State Increment no\_improvement\_count by 1.
                \EndIf
            \EndIf
        \EndIf
        \If{no\_improvement\_count $\geq$ patience}
            \State Terminate Training: Early stopping due to no improvement in MS-SSIM and FID scores for patience epochs.
            \State \textbf{break}
        \EndIf
    \EndFor
    \State End Training.
    \Return Loss values, Best MS-SSIM, and FID scores.
\end{algorithmic}
\end{algorithm}

\section{Results and Discussion}
\label{results_discussion}
\begin{figure}[ht!]
    \centering
    \includegraphics[width=0.45\textwidth]{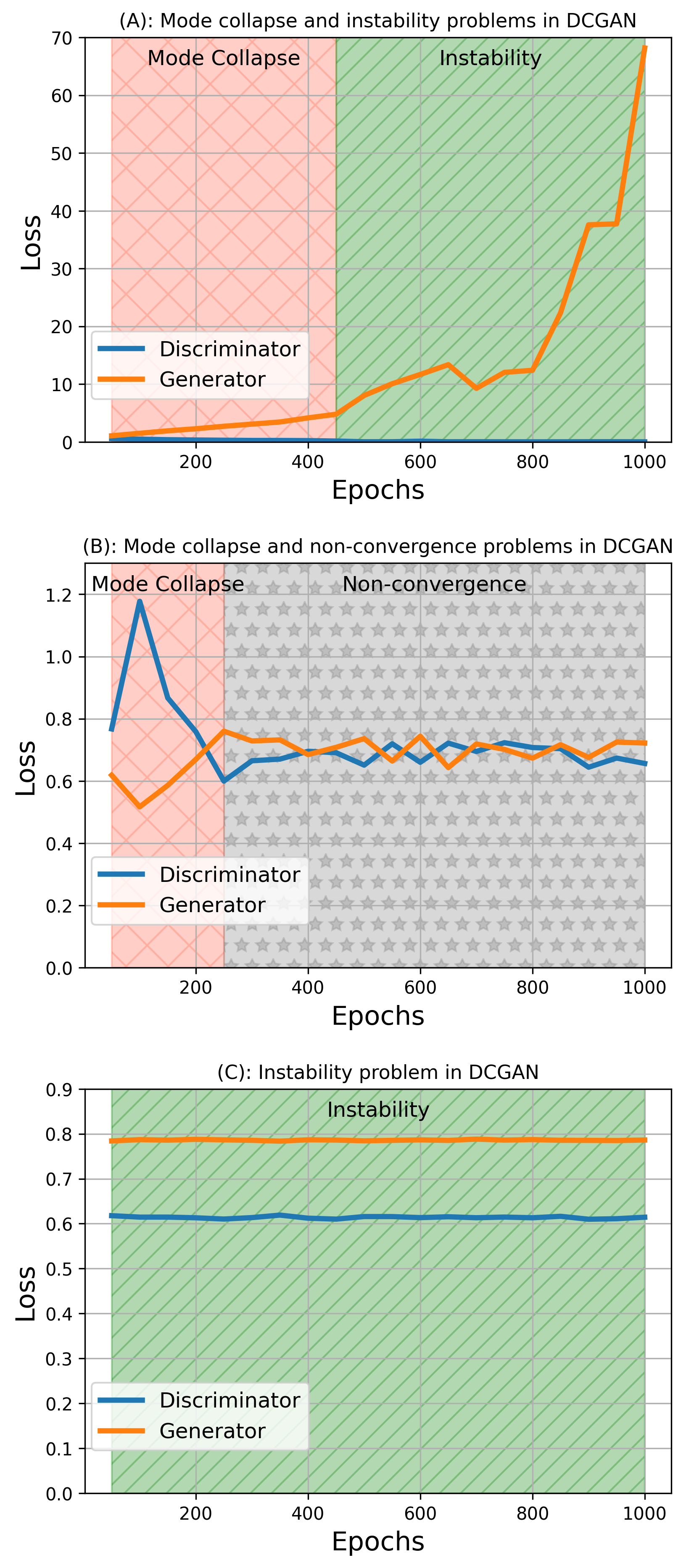}
    \caption{Quantitative analysis of loss behavior in DCGAN for identifying the mode collapse, non-convergence, and instability problems for synthetic COVID-19 X-ray images.}
    \label{fig:trainproblmloss}
\end{figure}

The loss values for the DCGAN training with LR 0.001, 0.002, and 0.003 are depicted in Fig. \ref{fig:trainproblmloss}. The results demonstrate that the correlation between generator and discriminator loss values significantly impacts the synthetic images.

As the discriminator loss value approaches zero, no meaningful gradient feedback is being backpropagated to the generator, increasing the generator loss (0-5) from epochs 0 to 450, indicating a mode collapse (See Fig. \ref{fig:trainproblmloss}(A)). Subsequently, the generator loss continues to increase (5-70) from epochs 450 to 1000 epochs without receiving meaningful feedback from the discriminator indicating the instability problem.

In Fig. \ref{fig:trainproblmloss}(B) the sharp increase/decrease in the generator and discriminator losses from epoch 0 to 250 indicates mode collapse. Subsequently, the non-convergence problem triggers oscillating loss values (0.6 to 0.75) from epoch 250 to 1000 due to the inconsistent gradient values propagated from the discriminator to the generator.

In Fig. \ref{fig:trainproblmloss}(C) the stable loss values for the discriminator (0.6 to 0.625) and the generator (0.775 to 0.8) indicate the instability problem throughout training, as there are small loss values with no improvements.

Table \ref{ganoverviewimplementeffect} depicts the relationship between the loss values and the visual quality of synthetic images. Identifying training problems via the quantitative measure of loss values is inconsistent and can be impacted by the type of loss, images, architecture, LR, batch size, and training epochs.
\begin{table}[htp!]
\centering
\caption{\textbf{Qualitative analysis of training problems in DCGAN using binary cross-entropy loss and visual quality of synthetic images for COVID-19 chest X-ray images.}}

\begin{tabular}{p{0.5cm}p{2.2cm}p{2.3cm}p{2.1cm}}
\toprule
\textbf{T.P} & \multicolumn{3}{|c}{\textbf{Qualitative Assessment}} \\
& \textbf{Generator Loss} & \textbf{Discriminator Loss} & \textbf{Synthetic Images} \\
\midrule

M.C & A sharp increase in loss or constant loss for longer epochs & A sharp decrease in loss or constant loss for longer epochs & \includegraphics[width=2cm, valign=c]{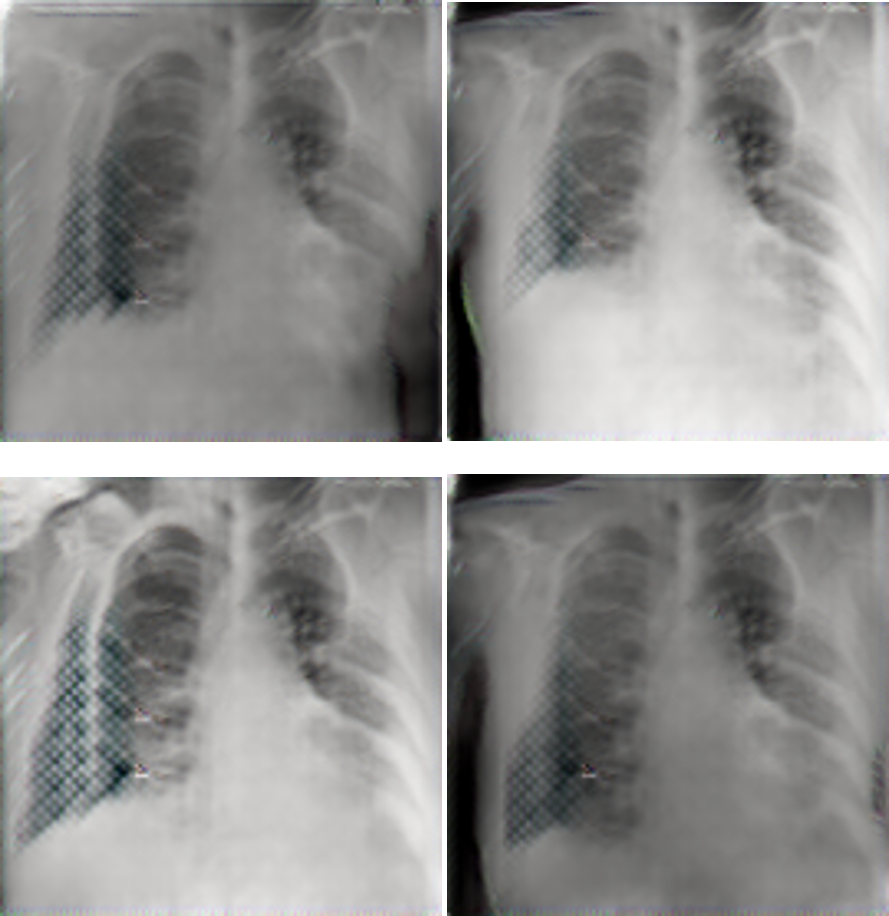} \\
N.C & Irregular Oscillations & Irregular Oscillations & \includegraphics[width=2cm, valign=c]{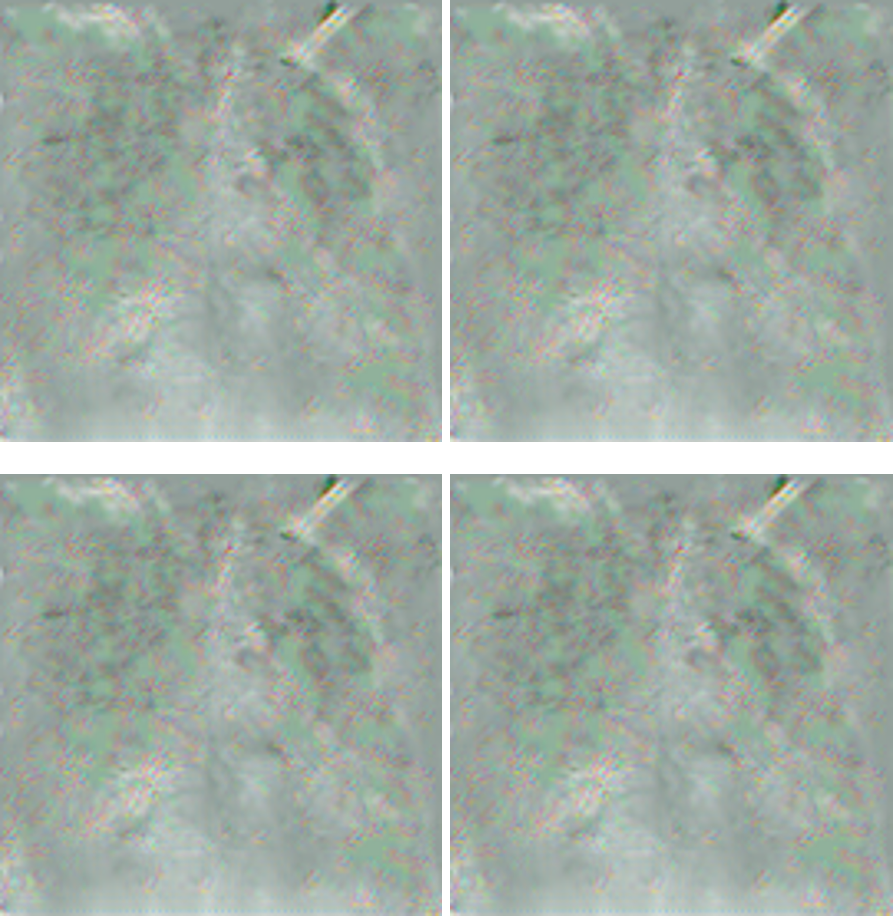} \\
Inst. & Abrupt jumps or drops or constant & Abrupt jumps or drops or constant & \includegraphics[width=2cm, valign=c]{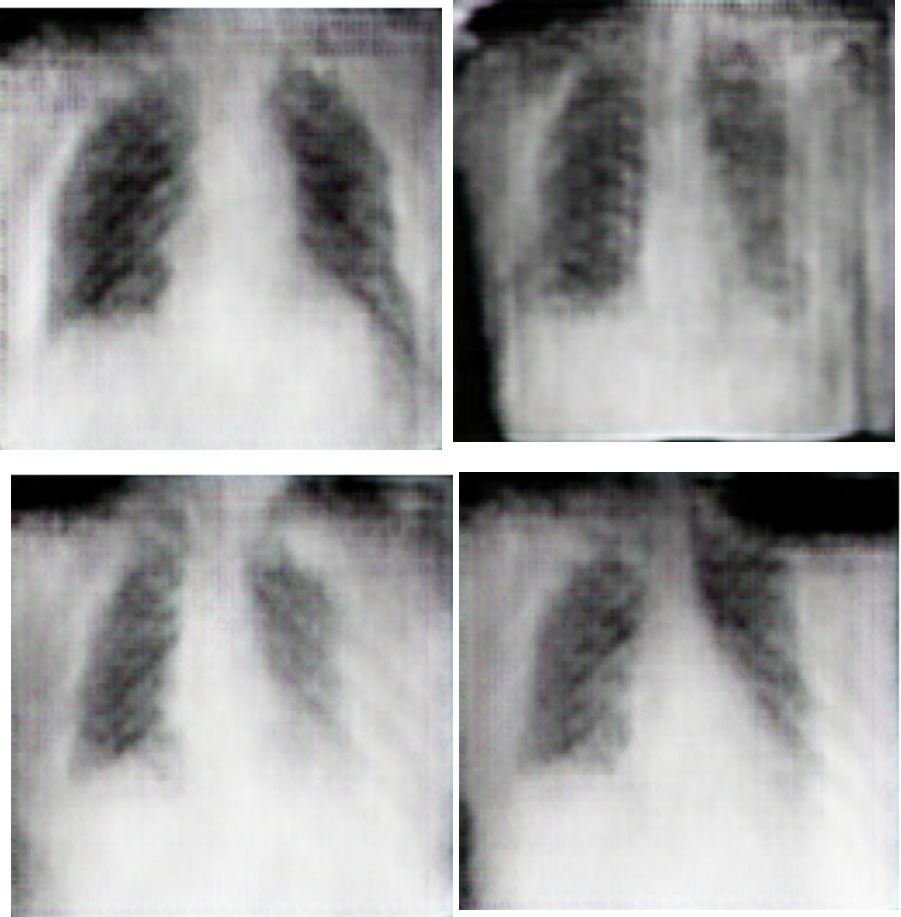} \\

\bottomrule
\multicolumn{4}{l}{Inst: Instability; M.C: Mode Collapse; N.C: Non-convergence} \\
\multicolumn{4}{l}{T.P: Training Problem} \\

\end{tabular}
\label{ganoverviewimplementeffect}
\end{table}

\begin{figure}[htp!]
    \centering
    \includegraphics[width=0.45\textwidth]{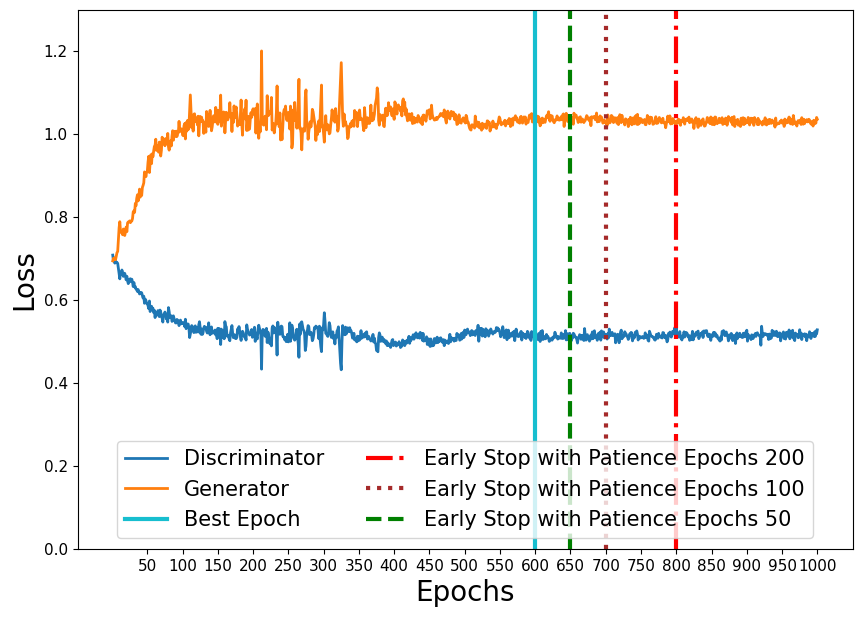}
    \caption{DCGAN training depicting early stopping points using loss values.}
    \label{fig:dcganlossPEps}
\end{figure}

Figure \ref{fig:dcganlossPEps} depicts DCGAN (LR=0.0001) experiencing consistent loss values for the generator and discriminator. The generator loss remains constant approaching 1 while the discriminator loss values remain constant approaching 0.5. In this DCGAN training, early stopping is implemented with loss values using different ranges of patience epochs as depicted in Fig \ref{fig:dcganlossPEps}. When the loss values of the generator and discriminator start behaving consistently in training, the training will be stopped after a fixed number of epochs indicating no more improvement in GANs. The patience of 200 epochs is significant as it allows more time to check for further improvement in the training of GANs. Notably, a stable training process is achieved when the generator loss becomes twice the discriminator loss, as measured by BCE loss in DCGAN. However, it is difficult to halt the training using these consistent loss values for the desired diversity and quality of synthetic images.

\begin{figure}[hbt!]
  \centering
  \begin{subfigure}[b]{0.45\textwidth}
    \centering
    \includegraphics[clip,width=\textwidth,height=5cm]{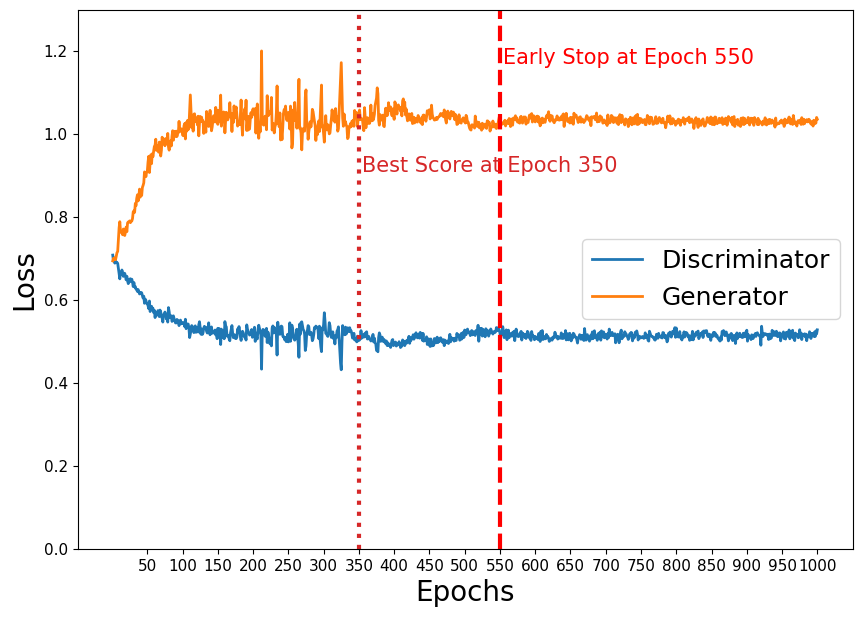}
    \caption{DCGAN loss values.}
  \end{subfigure}
  \hfill
  \begin{subfigure}[b]{0.45\textwidth}
    \centering
    \includegraphics[clip,width=\textwidth,height=5cm]{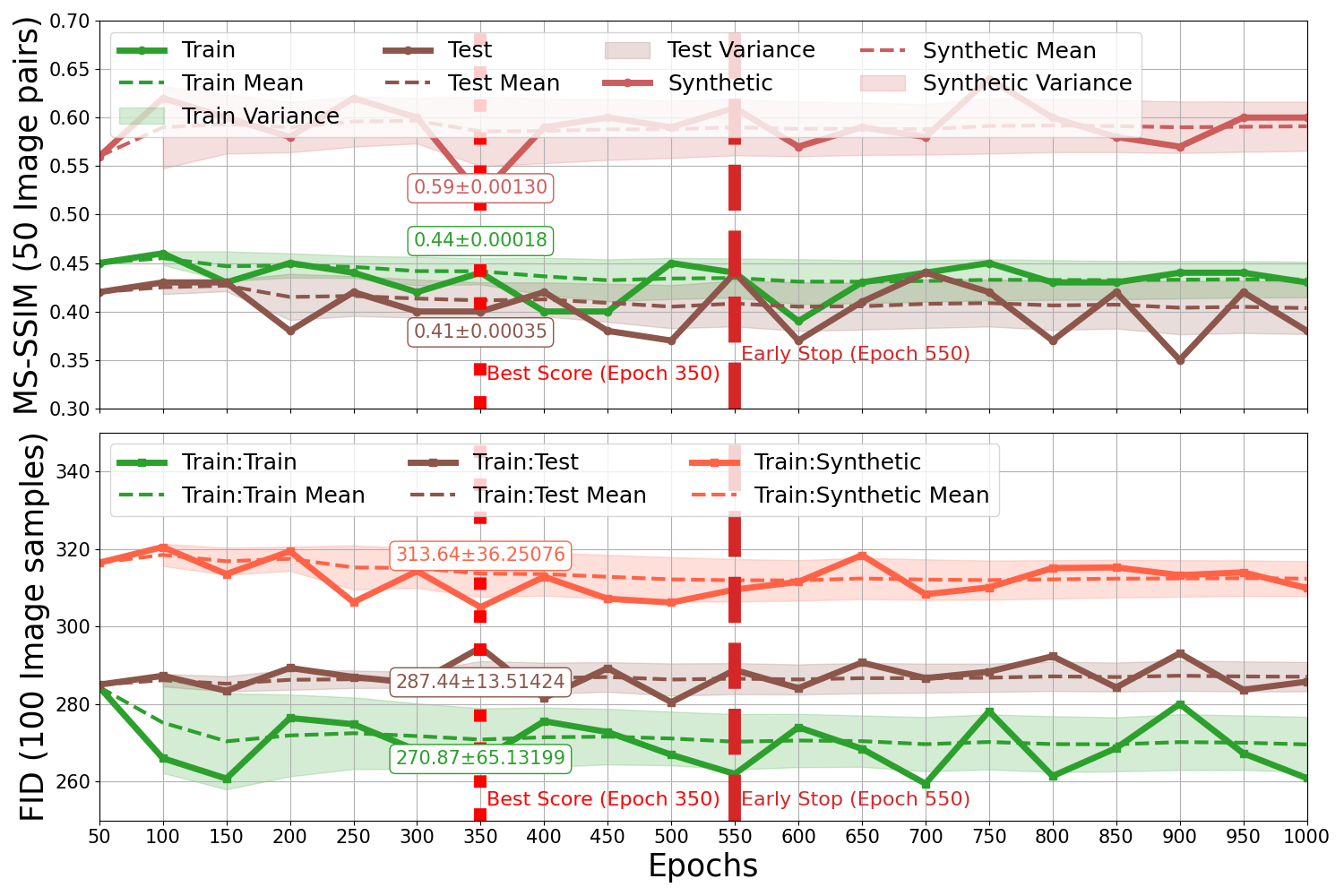}
    \caption{DCGAN MS-SSIM and FID scores.}
  \end{subfigure}
  \caption{DCGAN training depicting early stopping points for loss values combined with the MS-SSIM, and FID scores of synthetic images. The lower scores of MS-SSIM and FID indicate better diversity and quality of synthetic images.}
  \label{dcgan_data}
\end{figure}

Figure \ref{dcgan_data} depicts DCGAN (LR=0.0001) with early stopping criteria implemented using loss, MS-SSIM, and FID scores. The consistent loss values (See Fig. \ref{dcgan_data}(a)) result in the generation of diversified and high-quality images as indicated by the MS-SSIM and FID scores (See Fig. \ref{dcgan_data}(b)). Figure \ref{dcgan_data} indicates that when early stopping is implemented for a combination of loss values, MS-SSIM, and FID scores, the training can be stopped systematically to save time and computational resources as indicated in Table \ref{earlystoppoverview}. The training was halted in DCGAN at epoch 550 (patience 200 epochs) as the MS-SSIM and FID scores along with losses indicate the best values at epoch 350 (See Fig. \ref{dcgan_data}(b)) compared to train and test thresholds and no further improvements were detected in the subsequent 200 epochs.

\begin{figure}[hbt!]
  \centering
  \begin{subfigure}[b]{0.45\textwidth}
    \centering
    \includegraphics[clip,width=\textwidth,height=5cm]{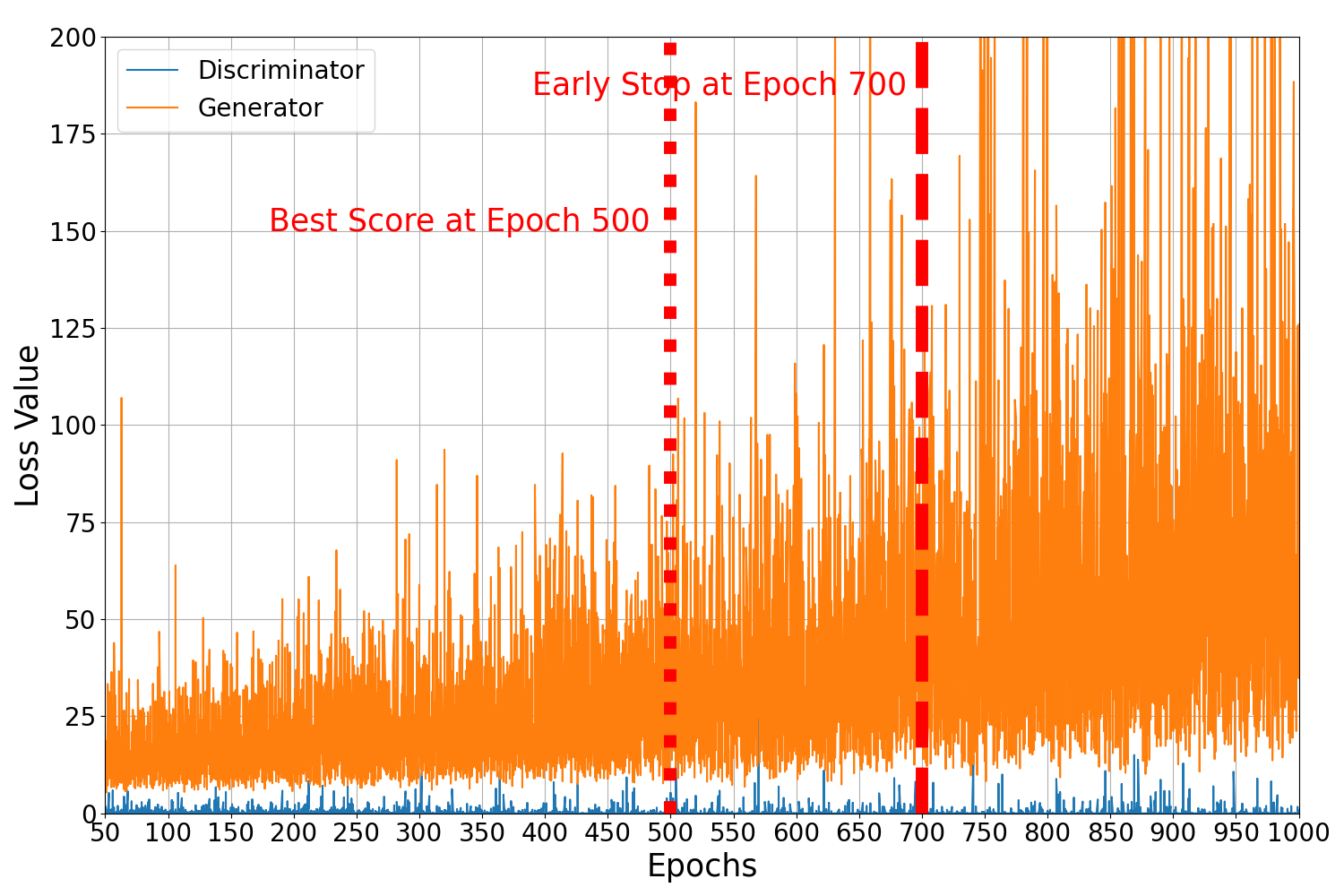}
    \caption{MSG-GAN loss values.}
  \end{subfigure}
  \hfill
  \begin{subfigure}[b]{0.45\textwidth}
    \centering
    \includegraphics[clip,width=\textwidth,height=5cm]{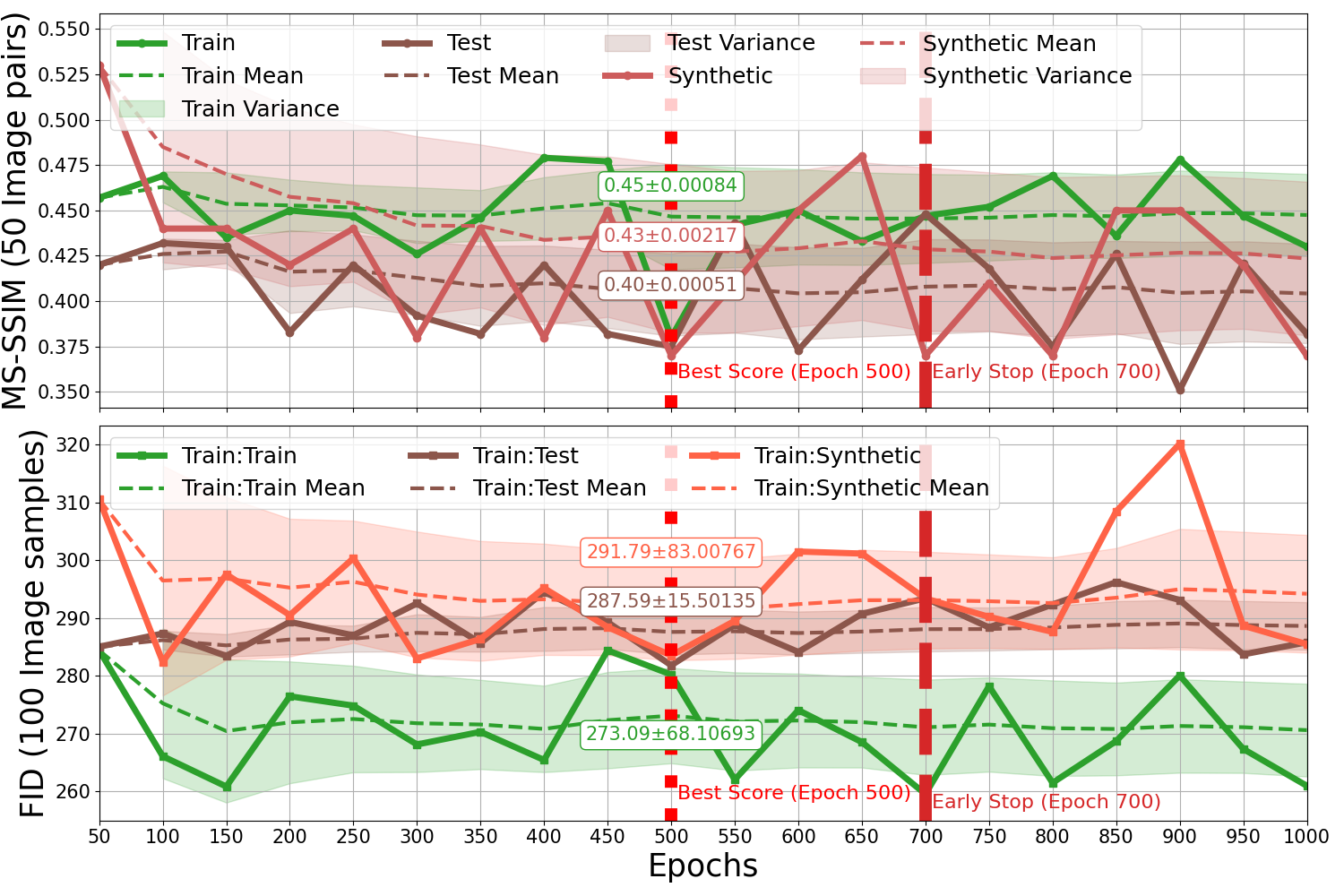}
    \caption{MSG-GAN MS-SSIM and FID scores.}
  \end{subfigure}
  \caption{MSG-GAN training depicting early stopping points for loss values combined with the MS-SSIM, and FID scores of synthetic images. The lower scores of MS-SSIM and FID indicate better diversity and quality of synthetic images.}
  \label{msg-gan_data}
\end{figure}

Figure \ref{msg-gan_data} depicts MSG-GAN with early stopping criteria implemented using loss, MS-SSIM, and FID scores. In MSG-GAN, epoch 500 indicates better scores of MS-SSIM and FID (See Fig. \ref{msg-gan_data}(b) while training was halted at epoch 700 with patience of 200 epochs indicating no further improvements in MS-SSIM and FID scores. There is no clear indication of loss values corresponding to the MS-SSIM and FID scores. The correlation between the loss values of the generator and the discriminator in MSG-GAN is different than in DCGAN. The reason is that the MSG-GAN is trained using the Relativistic Hinge loss indicating the occurrence of all three training problems throughout the training for 1000 epochs, as depicted in Fig. \ref{msg-gan_data}(a). The distinct behavior of MSG-GAN loss indicates the need for additional metrics in assessing the training problems for implementing the proposed early stopping criteria.

\begin{table}[htp!]
\centering
\caption{\textbf{A comparison of the number of epochs and computational time during the training of DCGAN and MSG-GAN with and without early stopping criterion.}}

\begin{tabular}{p{1.8cm}p{1.5cm}p{2cm}p{1.8cm}}
\toprule
\textbf{GANs} & \textbf{Max Epochs} & \textbf{Patience Epochs} & \textbf{Training Time} \\
\midrule

DCGAN & 1000 & n/a & 19 Hours \\
MSGAN & 1000 & n/a & 181 Hours \\
DCGAN (ES) & 550 & 200 & 10 Hours \\
MSGAN (ES) & 700 & 200 & 126 Hours \\

\bottomrule
\multicolumn{4}{l}{ES: Early Stopping; Max: Maximum Number} \\

\end{tabular}
\label{earlystoppoverview}
\end{table}
In this work, the MS-SSIM and FID scores are computed dynamically to analyze the range of values for train, test, and synthetic images. The diversity and quality of images using MS-SSIM and FID scores significantly vary across epochs for DCGAN and MSG-GAN. It indicates the significance of our approach as the literature lacks highlighting of this information.

This work is limited to a single dataset of COVID-19 chest X-ray images. The early stopping criterion and identification of training problems in the DCGAN and MSG-GAN will be investigated for alternate datasets such as pneumonia and lung X-ray images as part of future work.
\section{CONCLUSIONS}
The training of GANs is computationally expensive due to the complex architectures requiring high GPU memory and a longer training time. Furthermore, a GAN faces training problems such as mode collapse, non-convergence, and instability, affecting the diversity and quality of synthetic images desired in several application domains. The identification and quantification of these problems using loss and evaluation metrics are challenging. In this paper, DCGAN is adopted due to its requirements of few computational resources and less time taken in one training cycle to generate biomedical images. The training problems are addressed systematically by evaluating the loss values of the generator and discriminator. The early stopping criterion is proposed based on quantifying loss, MS-SSIM, and FID scores to save on computational resources and training time. Researchers can benefit from this early stopping criterion when training GANs using different loss functions and parameter settings. Synthetic images are generated at any epoch without focusing on loss values, diversity, or quality. Through early stopping, quantifying loss values agreed with MS-SSIM and FID scores will help to generate diversified and high-quality images while saving computational resources and training time. The patience in epochs is an arbitrary number that can be changed based on GAN training. A higher number of 200 is significant allowing GAN training to check for improvement. 




\end{document}